# Data-driven Insights for Informed Decision-Making: Applying LSTM Networks for Robust Electricity Forecasting in Libya

**Asma Agaal**
Artificial Intelligence Department, Faculty of Technical Sciences, Sabha, Libya
E-mail: asma.agaal@sebhau.edu.ly
ORCID iD: https://orcid.org/0000-0002-8687-2605

**Mansour Essgaer***
Artificial Intelligence Department, Faculty of Information Technology, Sebha University, Sabha, Libya
E-mail: man.essgaer@sebhau.edu.ly
ORCID iD: https://orcid.org/0000-0002-8447-5091
*Corresponding author

**Hend M. Farkash**
The Faculty of Electrical & Electronics Technology Benghazi, Libya
E-mail: mf_ceet@ceet.edu.ly
ORCID iD: https://orcid.org/0009-0009-5672-9873

**Zulaiha Ali Othman**
Research Center of Artificial Intelligent Technology, Faculty of Information Science and Technology, University Kebangsaan Malaysia, Malaysia, Selangor, Malaysia
E-mail: zao@ukm.edu.my
ORCID iD: https://orcid.org/0000-0002-4238-5266

**Abstract:** Accurate electricity forecasting is vital for grid stability and effective energy management, particularly in regions like Benghazi, Libya, which face frequent load shedding, generation deficits, and aging infrastructure. This study introduces a data-driven framework to forecast electricity load, generation, and deficits for 2025 using historical data from two distinct years: 2019 (an instability year) and 2023 (a stability year). Various time series models were employed, including Autoregressive Integrated Moving Average (ARIMA), seasonal ARIMA, dynamic regression ARIMA, extreme gradient boosting, simple exponential smoothing, and Long Short-Term Memory (LSTM) neural networks. Data preprocessing steps—such as missing value imputation, outlier smoothing, and logarithmic transformation—are applied to enhance data quality. Model performance was evaluated using metrics such as mean squared error, root mean squared error, mean absolute error, and mean absolute percentage error. LSTM outperformed other models, achieving the lowest mentioned metric values for forecasting load, generation, and deficits, demonstrating its ability to handle non-stationarity, seasonality, and extreme events. The study's key contribution is the development of an optimized LSTM framework tailored to North Benghazi's electricity patterns, incorporating a rich dataset and exogenous factors like temperature and humidity. These findings offer actionable insights for energy policymakers and grid operators, enabling proactive resource allocation, demand-side management, and enhanced grid resilience. The research highlights the potential of advanced machine learning techniques to address energy-forecasting challenges in resource-constrained regions, paving the way for a more reliable and sustainable electricity system.

**Index Terms:** Time Series Analysis, Electricity Forecasting, Load Forecasting, Generation Forecasting, Deficit Forecasting, Energy Management, North Benghazi Power Plant.

Data-driven Insights for Informed Decision-Making: Applying LSTM Networks for Robust Electricity Forecasting in Libya

## 1. Introduction

Electricity is a fundamental resource supporting modern society, making a reliable and efficient power supply essential. This requires accurate forecasting of critical variables such as load demand, generation capacity, and potential deficits [1]. However, the complexity of electricity systems presents significant challenges for accurate predictions, particularly with fluctuating demand patterns and the growing integration of intermittent renewable energy sources like solar and wind [2].

Reliable electricity forecasting is crucial for maintaining the stability and efficiency of power systems. By accurately predicting load fluctuations, system operators can effectively schedule generation resources, optimize transmission capacity, and mitigate the risk of blackouts or brownouts[3]. Additionally, accurate forecasts play a vital role in economic dispatch, enabling efficient trading and pricing of electricity based on supply and demand dynamics [4]. Accurate forecasting of renewable energy generation is also key for a sustainable energy future, maximizing utilization and reducing reliance on fossil fuels.

While advanced forecasting techniques are essential for managing electricity systems worldwide, their practical application often requires addressing region-specific challenges. In the case of Benghazi, Libya's second-largest city, unique circumstances such as political instability and infrastructure damage have exacerbated energy supply issues [5, 6], particularly during peak summer months [5]. Understanding and addressing these localized challenges is critical to implementing effective forecasting and energy management strategies. To address these issues, researchers have explored various solutions, including machine learning techniques for load forecasting [7] and the implementation of renewable energy sources, particularly solar power [6, 8], Despite Libya's substantial renewable energy potential, particularly in solar and wind energy, progress remains limited due to highly subsidized electricity tariffs and the absence of clear energy legislation [9, 10]. Understanding and addressing these localized challenges is critical to implementing effective forecasting and energy management strategies.

To address the complex forecasting challenges, advanced analytical techniques are needed to handle the complexities of electricity data. Time series analysis, a statistical methodology specifically designed to analyze data indexed by time, provides a powerful framework for this purpose [11]. Electricity data inherently exhibits temporal dependencies, with past values influencing future behavior [12]. Time series analysis techniques excel at capturing and modeling these dependencies, enabling the development of accurate and robust forecasting models [13].

This study leverages time series analysis to forecast key electricity variables for the North Benghazi Power Plant (NBPP) in eastern Libya, aiming to enhance operational efficiency and strategic planning. The research centers on predicting three critical and interrelated components: electricity load, electrical deficit estimation, and power generation. These elements are intrinsically connected—accurate load forecasting enables precise deficit estimation, which subsequently informs effective power generation strategies. To investigate these dependencies and identify the most suitable model for each forecasting task, we adopt a comparative approach, assessing the performance of three prominent time series models: Autoregressive Integrated Moving Average (ARIMA) [14], Seasonal ARIMA (SARIMA) [15], Dynamic Regression ARIMA (ARIMAX) [16], Simple Exponential Smoothing (SES) [17], Long Short-Term Memory (LSTM) [18], and Extreme Gradient Boosting (XGBoost) [19].

The primary objective of this study is to design a robust and accurate forecasting framework tailored to NBPP's unique electricity consumption and generation patterns. Unlike conventional approaches, which often rely on traditional statistical models or generic machine learning techniques, this research advantages advanced Long Short-Term Memory (LSTM) networks. These networks are well suited for capturing complex non-linear patterns, long-term temporal dependencies, and the inherent volatility of electricity data. By incorporating exogenous variables such as temperature and humidity, the proposed framework enhances forecasting accuracy and provides actionable insights for energy policymakers and grid operators.

While traditional methods like ARIMA, SARIMA, and SES are widely used, they often struggle with non-stationarity, seasonality, and extreme events common in regions with unstable power supplies. Machine learning techniques such as XGBoost offer flexibility but may lack the ability to model effectively long-term dependencies. The LSTM-based framework proposed in this study overcomes these limitations through its memory cells and gating mechanisms, which retain and utilize information over extended periods. However, the model's computational complexity and the need for large datasets for training may pose challenges in resource-constrained environments, and its reliance on historical data may hinder adaptability to sudden changes in grid infrastructure or consumption patterns.

This study contributes to the field by: (1) developing an optimized LSTM-based forecasting framework tailored to NBPP unique electricity patterns, (2) demonstrating the superiority of LSTM over traditional and machine learning models in handling complex, non-linear, and volatile electricity data, and (3) offering actionable recommendations for improving energy management and grid reliability in regions with similar supply challenges. By addressing these objectives, this research aims to contribute to a more reliable and sustainable electricity supply for NBPP and comparable regions globally.

This paper is structured as follows: section 2 presents a review of the state-of-the-art methods. Afterward, section 3 presents the proposed details framework employed. We analyze the experimental results in section 4. In final, section 5 presents a conclusion and recommendation of the paper.

## 2. Related Works

Accurately forecasting electricity variables such as load, generation, and deficits is crucial for grid stability and efficient energy management. In recent years, various time series analysis techniques, from traditional statistical models to advanced machine learning algorithms, have been employed to address these challenges. This section compares prominent studies, highlighting their contributions, limitations, and emerging trends in electricity forecasting, with a focus on the advantages of LSTM over traditional models like ARIMA, SARIMA, ARIMAX, SES, and XGBoost.

Traditional models such as ARIMA are widely used for load forecasting, primarily due to their ability to capture linear dependencies and seasonality [20]. However, ARIMA struggles with non-linear patterns, and its performance diminishes for longer forecasting horizons, as shown by [21]. In contrast, machine-learning models like XGBoost offer the flexibility to handle non-linearity and complex interactions between variables. For example, [22] demonstrated XGBoost's superior performance compared to ARIMA for load forecasting in volatile environments. However, XGBoost, like ARIMA, is limited in capturing long-term dependencies, as highlighted by [23]. Similarly, SES, though commonly used, fails to capture long-term trends and seasonality effectively, limiting its forecasting capabilities [17].

To overcome these limitations, deep learning techniques, particularly LSTM networks, have gained prominence. LSTM's unique architecture, with memory cells and gating mechanisms, enables it to capture long-term dependencies and non-linear patterns more effectively than traditional models. [24] demonstrated LSTM's superior performance over ARIMA and Support Vector Machines (SVM) for short-term load forecasting, owing to its ability to learn complex non-linear relationships and model temporal dependencies. Furthermore, [25] highlighted LSTM's success in long-term electricity price forecasting, particularly its ability to incorporate seasonality and exogenous factors like weather, giving it a clear edge over ARIMA and SES.

In the context of electricity generation forecasting, particularly for renewable sources like wind and solar, ARIMA has been applied but faces challenges due to the volatile and intermittent nature of these energy sources [26]. Machine learning techniques, including ANN and Support Vector Regression (SVR), have shown better performance by capturing non-linear relationships and complex dependencies in renewable generation forecasting [27]. For electricity deficit forecasting, which predicts the gap between supply and demand, ARIMA is used, but its limitations in handling non-linearity and long-term trends persist [28]. Machine learning models like ANN and SVR have demonstrated improved accuracy, as evidenced by [29], especially when exogenous factors like weather, economic, and demographic data are included [30].

Regardless of the chosen forecasting technique, dealing with the non-stationary nature of electricity data is critical. Preprocessing methods such as differencing, seasonal decomposition, and variance stabilization are commonly applied to ensure data suitability for forecasting models [31]. Model performance is evaluated using metrics such as mean absolute percentage error (MAPE), root mean squared error (RMSE), and R-squared to assess accuracy and fit [32,33].

LSTM's ability to capture complex non-linear patterns, long-term dependencies, and volatility sets it apart from traditional models like ARIMA/SARIMA/ARIMAX and machine learning techniques like XGBoost and SES. LSTM's memory cells allow it to retain information over long periods, making it highly effective for forecasting tasks where past values significantly impact future outcomes [34]. In contrast, ARIMA/SARIMA models, while useful for linear patterns, are constrained by their reliance on fixed observation windows and linear assumptions, making them ineffective for modeling long-term trends and volatility [35]. SES, with its simple SES mechanism, fails to capture long-term dependencies, while XGBoost, despite being robust to non-linearity, cannot model temporal relationships effectively [36].

Moreover, LSTM excels at handling volatility and extreme events, such as sudden spikes in electricity demand, whereas traditional methods like ARIMA/SARIMA struggle due to their reliance on linear assumptions and fixed windows [37,38]. SES, while sensitive to outliers, often produces inaccurate forecasts in volatile situations. Though XGBoost is noise-robust, it lacks the temporal modeling capability to manage sudden changes effectively [39-41]. Additionally, LSTM's ability to integrate exogenous variables, such as temperature and humidity, further improves its forecasting precision by capturing the complex interactions between these external factors and the electricity data [42] [43]. ARIMAX and SES, however, rely on linear assumptions and cannot model these complex interactions adequately [44,45].

Despite LSTM's success in various forecasting applications, it has not been widely applied to forecast electricity load, generation, and deficits for specific regions, such as in NBPP, Libya. The unique electricity patterns and dynamics of this region, influenced by local climate conditions, socioeconomic factors, and infrastructure limitations, necessitate a tailored forecasting model. This study aims to address this gap by developing an LSTM-based forecasting model designed for the NBPP context, providing insights for improved energy management and grid reliability in the region.

## 3. Proposed Approach

Our proposed framework for forecasting electricity load, deficit, and generation in NBPP follows these key stages, also illustrated in Fig. 1:

- *Data Acquisition:*
  - Collection of historical electricity data from 2019 (instability) and 2023 (stability).

- Inclusion of relevant variables: electricity load, generation, deficit, temperature, and humidity.
- *Data Preprocessing:*
  - Imputation of missing values using linear interpolation.
  - Handling of outliers with a Savitzky-Golay filter.
  - Variance stabilization through logarithmic transformation.
- *Model Development*:
  - Evaluation of traditional statistical models, including ARIMA, SARIMA, and ARIMAX.
  - Implementation of advanced machine learning models, such as LSTM, XGBoost, and SES.
  - Optimization of the LSTM model through grid search and 5-fold cross-validation.
  - Incorporation of exogenous variables (temperature, humidity) into both ARIMAX and LSTM models.
- *Performance Evaluation*:
  - Comprehensive assessment using performance metrics: MSE, RMSE, MAE, and MAPE.
  - Comparison of model performance to identify the most accurate forecasting approach.

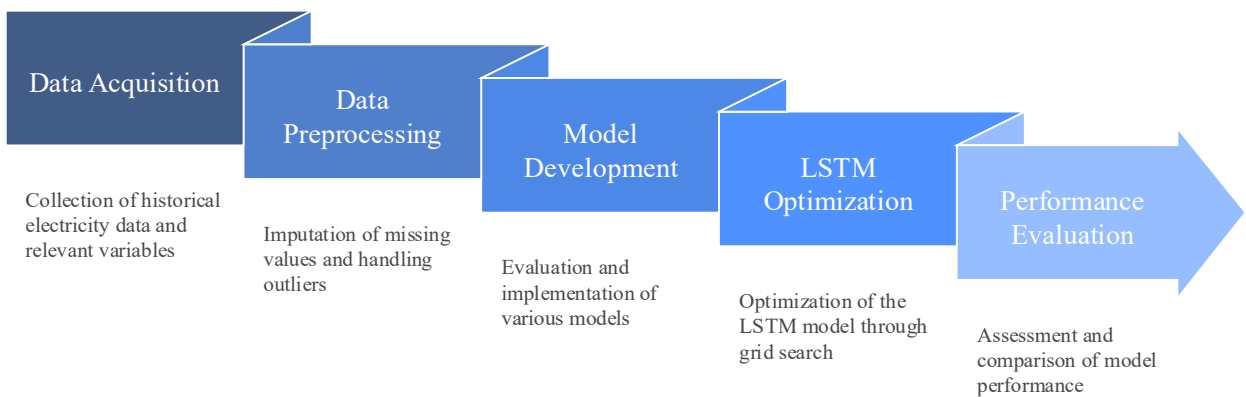

Fig.1. The proposed framework

### 3.1. Data Acquisition

To develop robust forecasting models, we collected a comprehensive dataset from the NBPP, located in the Eastern Region, called the 220-control area. The dataset includes daily observations of key electricity variables for the two distinct years 2019 and 2023. These years were selected to represent contrasting system behaviors: 2019, marked by instability with frequent load shedding and generation deficits, and 2023, characterized by relatively stable electricity conditions.

Table 1 summarizes the dataset's characteristics, providing an overview of the features variable. It distinguishes between the target variables (load, deficit, and generation) and additional features such as humidity, temperature, and date. These additional variables were analyzed for their potential influence on electricity patterns, thereby improving the forecasting models' predictive accuracy.

Table 1. Dataset characteristics

| Features | Description | Units | Target Variable (Yes/No) |
|---|---|---|---|
| Days | Unit of time for each observation (daily data) | Day | No |
| Hours | Hour of the day when data was recorded | Hour | No |
| Years | Year of data collection (2019, 2023) | Year | No |
| Generation | Electric Power Generated | Megawatt-hours (MWh) | Yes |
| Loads | Power Demanded by consumers | Megawatt-hours (MWh) | Yes |
| Deficit Amount | Power Shortage due to insufficient generation | Megawatt-hours (MWh) | Yes |
| Temperature | Ambient Temperature | Degrees Celsius (°C) | No |
| Humidity | Relative Atmospheric Humidity | Percentage (%) | No |

### 3.2. Data Preprocessing

Before conducting the analysis, the raw data underwent preprocessing to address missing values, outliers, and variance issues, ensuring its readiness for time series modeling. The key preprocessing steps included:

- Missing Value Imputation: Missing data points in the time series estimated using linear interpolation. This method ensured that the temporal structure of the dataset was preserved by calculating values based on neighboring data[46].
- Outlier Handling: a Savitzky-Golay filter [47] was applied to smooth the time series and minimize the influence of outliers. By fitting a local polynomial regression to the data, extreme values were reduced while long-term trends were retained.
- Logarithmic Transformation: A logarithmic transformation is applied to stabilize the variance of the time series. This process compressed large values, expanded smaller ones, and ensured compatibility with forecasting models like ARIMA, which assumes constant variance [48].

### 3.3. Autoregressive Integrated Moving Average

The Autoregressive Integrated Moving Average (ARIMA) model is a widely used statistical method for time series forecasting. ARIMA models effectively capture temporal dependencies by combining three core components [14]:

- Autoregressive (AR): This component captures the relationship between the current value of the time series and its past values. The order of the AR component is represented by $p$.
- Integrated (I): This component models the relationship between past forecast errors and the current value. The order of the MA component is indicated by $q$.
- Moving Average (MA): This component models the relationship between past forecast errors and the current value. The order of the MA component is denoted by q.

The equation (1) for the $ARIMA(p,d,q)$ model can be written as [49]:

$$\phi_P(B)(1-B)^d y_t = \theta_q(B)\epsilon_t \tag{1}$$

$y_t$ The time series at a time $t$,
$B$ the backshift operator (i.e., $B_{yt} = y_{t-1}$),
$\phi_p(B)$ represents the AR terms of order $p$ and can be written as $\phi_p(B) = 1 - \phi_1 B - \phi_2 B^2 - \cdots - \phi_p B^p$,
$\theta_p(B)$ represents the MA terms of order $q$ and can be written as $\theta_q(B) = 1 + \theta_1 B + \theta_2 B^2 + \cdots + \theta_q B^q$,
$(1-B)^d$ represents the differencing operator applied $d$ times to make the series stationary,
$\epsilon_t$ the white noise error term at a time $t$.

The complete ARIMA model is represented as ARIMA.$(p,d,q)$. Determining the optimal combination of $(p,d,q)$ Is crucial for accurate forecasting, which automatically identifies the optimal parameters by minimizing information criteria, such as the Akaike Information Criterion (AIC) or Bayesian Information Criterion (BIC) [50].

### 3.4. Seasonal Autoregressive Integrated Moving Average

When time series data exhibits recurring seasonal patterns, the Seasonal ARIMA (SARIMA) model extends the standard ARIMA framework by incorporating additional components to capture periodic fluctuations alongside the overall trend. SARIMA enhances the model's ability to account for seasonality by including seasonal counterparts to the standard AR, I, and MA components [51]:

- Seasonal Autoregressive (SAR): Captures the relationship between past values of the time series that occur at identical seasonal intervals.
- Seasonal Integrated (SI): Applies seasonal differencing to address stationarity in data with recurring seasonal patterns.
- Seasonal Moving Average (SMA): Models the relationship between past forecast errors at corresponding seasonal intervals and the current value.

Equation (2) general representation of the SARIMA model, combining both non-seasonal and seasonal components [52]:

$$\Phi_P(B^s)\phi_p(1-B)^d(1-B^s)^D y_t = \Theta_Q(B^s)\theta_q(B)\epsilon_t \tag{2}$$

$y_t$ is the time series at a time $t$,
$B$ is the backshift operator (i.e., $\boldsymbol{B_{yt} = y_{t-1}}$),
$\phi_p(B)$ represents the non-seasonal (AR) terms of order $p$,
$\theta_q(B)$ represents the non-seasonal (MA) terms of order $q$,
$\Phi_P(B^s)$ represents the seasonal AR terms of order $P$,
$\theta_Q(B^s)$ represents the seasonal SMA terms of order Q,
$d$ is the non-seasonal differencing order, and $D$ is the seasonal differencing order,
$s$ The length of the seasonal cycle (e.g., 12 for monthly data).

$\epsilon_t$ Error term at a time $t$.

This equation (2) captures both short-term and seasonal trends in the time series, making it suitable for datasets with seasonal patterns, such as electricity usage influenced by factors like temperature and holidays.

*3.5. Dynamic Regression Autoregressive Integrated Moving Average*

Dynamic Regression ARIMA, also known as ARIMAX (ARIMA with exogenous variables), extends the traditional ARIMA model by incorporating external (temperature, humidity, holidays) variables that may influence the time series being modeled [53]. These exogenous variables are denoted as $x_t$, can improve forecasting by accounting for factors outside the historical time series. Equation (3) illustrates the ARIMAX model [54]:

$$\phi^d_{p(B)(1-B)} y_t = \theta_{q(B)\epsilon_t} + \beta_0 + \beta_1 x_{\{1,t\}} + \beta_2 x_{\{2,t\}} + \ldots + \beta_k x_{\{k,t\}} \quad (3)$$

$y_t$ Is the time series at a time $t$,
$B$ is the backshift operator (i.e., $\boldsymbol{B_{yt} = y_{t-1}}$),
$\phi_p(B)$ represents the non-seasonal autoregressive (AR) terms of order $p$,
$\theta_q(B)$ represents the non-seasonal moving average (MA) terms of order $q$,
$(1-B)^d$ represents the differencing operator applied $d$ times to make the series stationary,
$\epsilon_t$ the white noise error term at a time $t$.
$x_{1,t}, x_{2,t}, \ldots, x_{k,t}$ are the exogenous (external) variables at a time $t$,
$B_0, B_1, \ldots, B_K$ the coefficients for the exogenous variables.

*3.6. Long Short-Term Memory Neural Network*

Long Short-Term Memory (LSTM) is a specialized type of recurrent neural network (RNN) designed to handle sequence prediction tasks by effectively capturing long-term dependencies. LSTMs address the vanishing gradient problem that traditional RNNs face, making them particularly well-suited for time series forecasting [55].

The architecture of an LSTM consists of three primary gates: the forget gate, the input gate, and the output gate. These gates control the flow of information within the network, determining which information should be retained, updated, or output based on the input data [19]. The following are detailed descriptions of these gates:

**Forget Gate:** The forget gate as shown in equation (4) determines which information from the previous cell state $C_t-1$ should discarded:

$$f_t = \sigma(W_f \cdot [h_{\{t-1\}}, x_t] + b_f) \quad (4)$$

$f_t$ The forget gate output,
$W_f$ and $f_t$ are the weight matrix and bias for the forget gate,
$h_{t-1}$ the hidden state from the previous time step,
$x_t$ the input at the current time step,
$\sigma$ Sigmoid activation function.

**Input Gate:** the input gate controls the new information to be stored in the cell state, as described by the following equations (5) and (6), Equation (5) models the input gate's output, which determines how much of the new information should be added:

$$i_t = \sigma(W_i \cdot [h_{\{t-1\}}, x_t] + b_i) \quad (5)$$

Equation (6) represents the candidate cell state $C_t$ which is the new information that could be added to the cell state:

$$C_t = \tanh(W_c \cdot [h_{\{t-1\}}, x_t] + b_c) \quad (6)$$

$i_t$ the input gate output,
$C_t$ the candidate's cell state (the new information that could be added),
$W_c$ the weight matrix for the candidate cell stat.
$b_c$ the bias term.
*tanh* the hyperbolic tangent activation function.

**Cell State Update:** The cell state is updated by combining the outputs of the forget gate and input gate, as described in the following equation (7).

Data-driven Insights for Informed Decision-Making: Applying LSTM Networks for Robust Electricity Forecasting in Libya

$$C_t = f_t \cdot C_{t-1} + i_t \cdot C_t \tag{7}$$

$CtC_t$ the updated cell state,
$Ct-1C_{t-1}$ the previous cell state.

**Output Gate**: The output gate determines the hidden state $h_t$ and decides what to output from the current cell state. This is represented in the following equations (8) and (9):

$$O_t = \sigma(Wo \cdot [h_{t-1}, X_t] + b_o) \tag{8}$$

$$h_t = o_t . \tan h(C_t) \tag{9}$$

$OtO_t$ is the output gate value,
$hth_t$ is the new hidden state (which is also the output of the LSTM at the time **t**)
$Wo$ and $boWo$ are the weight matrix and bias for the output gate.

This gated architecture enables LSTMs to selectively retain or discard information over extended sequences, allowing them to capture complex temporal dependencies that other models, such as ARIMA, might miss.

*3.7. Extreme Gradient Boosting*

Extreme Gradient Boosting (XGBoost) is built on gradient boosting, a technique that sequentially constructs an ensemble of decision trees to minimize errors from previous models [56]. The core of XGBoost lies in its loss function, which quantifies the difference between the actual and predicted values, as shown in equation (10):

$$Loss = \sum_{i=1}^{n} L(y_i, \hat{y}_i) \tag{10}$$

$n$ is the number of data points.
$L(y_i, \hat{y}_i)$ is a function that computes the error between the actual value $y_i$ and the predicted value $\hat{y}_i$. For example, in regression, this could be Mean Squared Error (MSE), and in classification, it could be Log Loss.

To prevent overfitting, XGBoost incorporates a regularization term that penalizes model complexity [45], as shown in equation (11):

$$\Omega(f_t) = \gamma T + \frac{1}{2}\lambda \sum_j w_j^2 \tag{11}$$

T is the number of leaves (or terminal nodes) in the tree.
γ is a parameter that controls the penalty on the number of leaves, discouraging over-complex trees.
$w_j$ represents the weight of the j-th leaf (how much contribution each leaf makes to the prediction).
λ is a parameter that controls the penalty on the leaf weights (larger values shrink the weights, leading to simpler models)

In each iteration, the model updates predictions by adding a new tree, as shown in equation (12):

$$\hat{y}_i^{(i+1)} = \hat{y}_i^{(t)} + \eta f_t(x_i) \tag{12}$$

Where η is the learning rate and $f_t(x_i)$ is the prediction of the new tree. The final prediction is the sum of all tree predictions, as shown in equation (13):

$$\hat{y}_i = \sum_{t=1}^{T} f_t(x_i) \tag{13}$$

$T$ is the total number of trees.
$f_t(x_i)$ is the prediction from the t-th tree for input $x_i$.

*3.8. Simple Exponential Smoothing*

Simple Exponential Smoothing (SES) is a widely used time series forecasting method that employs weighted averages of past observations to make predictions. The technique assigns exponentially decreasing weights to past data, with more recent observations having a greater influence on the forecast compared to older ones. SES is particularly effective for forecasting load data in time series with no strong trends or seasonal patterns [57].
The forecast $\hat{y}_{t+1}$ for the next period, *t+1* is as shown in equation (14):

$$\hat{y}_{t+1} = \alpha y_t + (1-\alpha)\hat{y}_t \tag{14}$$

- $\hat{y}_{t+1}$ is the forecast for the next period.
- α is the smoothing factor, 0<α<10, which controls the rate of decay of the weights for past observations.
- $y_t$ is the actual observation at time t.
- $\hat{y}_t$ is the forecast for period t.

### 3.9. Model Performance Improvement Techniques

*A. Auto Arima to Improve the ARIMA Models*

Auto ARIMA is a statistical method for time series forecasting that automates the process of determining the optimal parameters (p, d, q) for an ARIMA model. Instead of manually identifying these parameters through autocorrelation and partial autocorrelation plots, Auto ARIMA uses algorithms—typically based on information criteria like AICc or BIC—to iteratively search for the best combination of (p, d, q) values that minimize forecasting errors. However, Auto ARIMA may not always find the absolute best model, and its effectiveness depends on the specific characteristics of the time series data [58].

*B. Grid Search Technique for Improve the LSTM and XGBoost*

Grid search is a hyper-parameter tuning technique in machine learning used to find the optimal combination of hyper-parameters for a model. Hyper-parameters are the settings that control the learning process, such as the learning rate in a neural network. Grid search systematically explores all possible combinations of hyperparameters within specified ranges, evaluating the model's performance for each combination [59].

### 3.10. Model Evaluation

The evaluation metrics for time series models are classified into two categories. The first category focuses on assessing the stability of the time series, while the second category evaluates the accuracy and performance of the forecasting models.

*The Stability of the Time Series*

Augmented-Dickey-Fuller (ADF) and p-values: The ADF test is a formal test for stationarity. It extends the basic Dickey-Fuller test by including lagged differences of the series to account for higher-order autocorrelations [60]. Equation (15) shows that the regression equation ADF is:

$$\Delta Y_t = \alpha + \beta t + \gamma Y_{t-1} + \sum_{i=1}^{p} \sigma_i \Delta Y_{t-1} + \epsilon_t \tag{15}$$

$Y_t$ is the value of the series at time t,
$\Delta Y_t = Y_t - Y_{t-1}$ is the first difference of the series,
α is the constant term,
$\beta t$ represents the time trend (optional),
γ is the coefficient for the lagged level of the series, which indicates the presence of a unit root,
$\sigma_i$ are coefficients for the lagged differences of the series,
p is the number of lags of the different terms,
$\epsilon_t$ is the error term.

The null hypothesis is that γ=0 (indicating a unit root, non-stationary), while the alternative hypothesis is that γ<0 (stationary) [61].

Autocorrelation Function (ACF): The ACF is used to check the correlation between the time series and its lagged values. For a stationary time series, the autocorrelations should decay rapidly to zero [62].

The autocorrelation function for lag *k* given by equation (16):

$$\rho(k) = \frac{\sum_{t=k+1}^{T}(Y_t - \bar{Y})(Y_{t-k} - \bar{Y})}{\sum_{t=1}^{T}(Y_t - \bar{Y})^2} \tag{16}$$

$Y_t$ is the value of the series at time t.
$\bar{Y}$ is the mean of the series,
k is the lag,
T is the total number of observations.

For a stationary time series, *ρ(k)* should approach zero as *k* increases.

Partial Autocorrelation Function (PACF): The PACF measures the correlation between a time series and its lagged values after accounting for the effects of intermediate lags [63].

As shown in equation (17), the PACF at lag k, $\alpha(k)$, is the coefficient of $Y_{t-k}$ in the regression:

$$Y_t = \alpha 1 Y_{t-1} + \alpha 2 Y_{t-2} + \cdots + \alpha k Y_{t-k} + \epsilon t \tag{17}$$

Where $\epsilon t$ is the error term. The PACF should drop to zero after a few significant lags in a stationary series.

*The Accuracy and Performance of Models*

Mean Squared Error (MSE): Measures the average of the squared differences between predicted and actual values, emphasizing larger errors [64]. The calculation is expressed as equation (18):

$$MSE = \frac{1}{n}\sum_{i=1}^{n}(Y_i - Y_i^\sim)^2 \tag{18}$$

where $Y\_i$ is the actual value, $Y\_i^\wedge\sim$ is the predicted value, and n is the number of observations [65].

Mean Absolute Error (MAE): This represents the average of the absolute differences between predicted and actual values, providing an easily interpretable measure of error [66]. It is calculated as the equation (19):

$$RMSE = \sqrt{\frac{1}{n}\sum_{i=1}^{n}(Y_i - Y_i^\sim)^2} \tag{19}$$

Mean Absolute Percentage Error (MAPE): Measures prediction accuracy as a percentage by averaging the absolute percentage errors between predicted and actual values [67]. It is calculated as the equation (20):

$$\text{MAPE} = \frac{100\%}{n}\sum_{i=1}^{n}\left|\frac{Y_i - Y_i^\sim}{Y_i}\right| \tag{20}$$

Mean Absolute Percentage Accuracy (MAPA): Provides a measure of accuracy by subtracting MAPE from 100%, indicating how close predictions are to the actual values [68]. The following equation (21) is used to determine the $MAPA$:

$$\text{MAPA} = \text{MAPE} \times 100\% \tag{21}$$

## 4. Result and Discussion

In this section, we present the analysis of the electricity load, generation, and deficit data, as well as the performance of various forecasting models, including ARIMA, XGBoost, and LSTM. The findings are discussed in detail, covering descriptive statistics, trend and seasonality analysis, model fitting, and comparative performance evaluation across all models. The experiential details are outlined as follows:

- *Overview of Data Analysis:*
  - Descriptive statistics for electricity load, generation, and deficit.
  - Identification of outliers and analysis of data distribution.

- *Heatmap and Relationship Analysis:*
  - Visualization of relationships and seasonal patterns over the past two years.
  - Insights into improvements in generation capacity.

- *Trend and Seasonality Analysis:*
  - Examination of system stability and seasonal fluctuations.

- *Model Fitting:*
  - Performance comparison of ARIMA, XGBoost, and LSTM models.

- *Comparative Model Analysis:*
  - Summary of LSTM's ability to capture complex, non-linear relationships and seasonal patterns.
  - Emphasis on LSTM's effectiveness for predicting load, generation, and deficit.

*Exploration Data Analysis*

Before building the models, we performed an extensive EDA to understand the electricity load, deficit, and generation data. This step was essential for identifying key patterns, assessing the time series forecasting suitability, detecting anomalies, and guiding data preprocessing and model selection. This section presents the main insights from our EDA, including descriptive statistics, time series visualizations, and stationarity testing.

Data-driven Insights for Informed Decision-Making: Applying LSTM Networks for Robust Electricity Forecasting in Libya

## A. Descriptive Statistics

Table 2 presents the descriptive statistics of the electricity data, offering insights into the central tendency and dispersion of key features. The data, with 729 daily observations in 2019 and 2023, reveals the following:

- *Generation and Load:* Average generation (1441.67 MWh) slightly exceeded average load (1426.04 MWh), suggesting a generally balanced system. However, substantial standard deviations (213.90 MWh and 233.29 MWh, respectively) highlight significant daily fluctuations in both generation and demand, emphasizing the dynamic nature of electricity supply and consumption.
- *Deficit:* The average daily deficit of 41.12 MWh, with a high standard deviation of 85.15 MWh, points to potential shortfalls on certain days. This variability underscores the need for accurate forecasting to manage supply shortages effectively.
- *Climatic Context:* Average temperatures of 24.28°C and humidity levels of 84% suggest a warm and humid climate, likely influencing electricity consumption patterns and contributing to demand fluctuations.

Table 2. Descriptive statistics of collected features

| Features | Mean | Std. Dev. | Min | 25th % | Median | 75th % | Max |
|---|---|---|---|---|---|---|---|
| Day | 15.72 | 8.80 | 1 | 8 | 16 | 23 | 31 |
| Month | 6.53 | 3.45 | 1 | 4 | 7 | 10 | 12 |
| Year | 2021.0 | 2.00 | 2019 | 2019 | 2021 | 2023 | 2023 |
| Generation (MWh) | 1441.9 | 213.86 | 567 | 1279.50 | 1410.50 | 1593.75 | 2063 |
| Loads (MWh) | 1426.2 | 233.21 | 200 | 1268.00 | 1405.00 | 1565.50 | 2106 |
| Deficit (MWh) | 41.12 | 85.15 | 0 | 0 | 0 | 51 | 785 |
| Temperature (°C) | 24.28 | 6.64 | 8 | 19 | 25 | 30 | 37 |
| Humidity (%) | 83.92 | 3.29 | 13 | 63 | 73 | 80 | 88 |

## B. Features Distribution and Outliers of Load, Generation, and Deficit

To understand the characteristics of individual features and identify potential anomalies, we examined the distribution of electricity load, generation, and deficit.

- *Electricity Load:* The load distribution, shown in Fig. 2, is slightly right-skewed, with most values ranging between 1000 kW and 1500 kW. Fewer instances exceed 1750 kW, and the average (around 1400 kW) is slightly higher than the median. The box plot highlights outliers exceeding 2000 kW, with some approaching 2100 kW. These occasional spikes in demand should considered in system planning and forecasting.

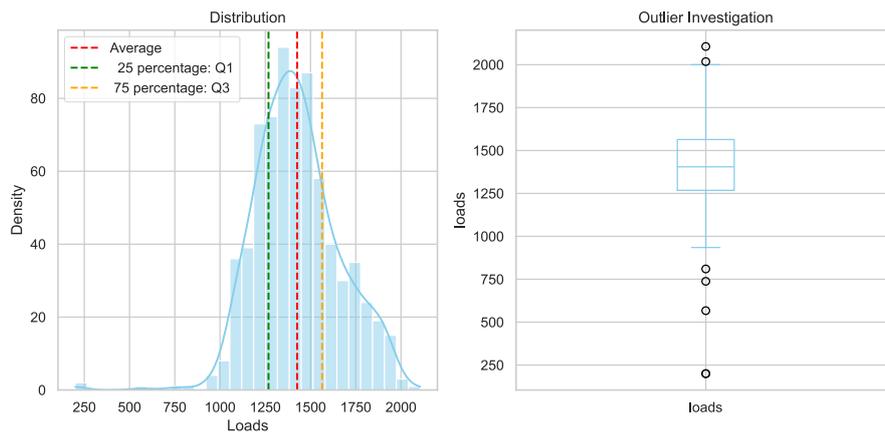

Fig.2. Load distribution and outlier investigation

- *Electricity Generation:* Shown in Fig. 3, generation follows a slightly right-skewed distribution. The average is around 1450 kW, with the interquartile range (IQR) between 1270 kW and 1600 kW, indicating variability. The box plot highlights three potential outliers below 800 kW, possibly due to planned maintenance or outages, which require further examination.
- *Electricity Deficit:* Fig. 4 shows a highly right-skewed distribution of deficits, with frequent small deficits and occasional large shortfalls. The average deficit is around 30 kW, but this is skewed by infrequent large deficits. The 25th percentile is close to zero, and the 75th percentile is below 50 kW, indicating minimal to no deficit most of the time.

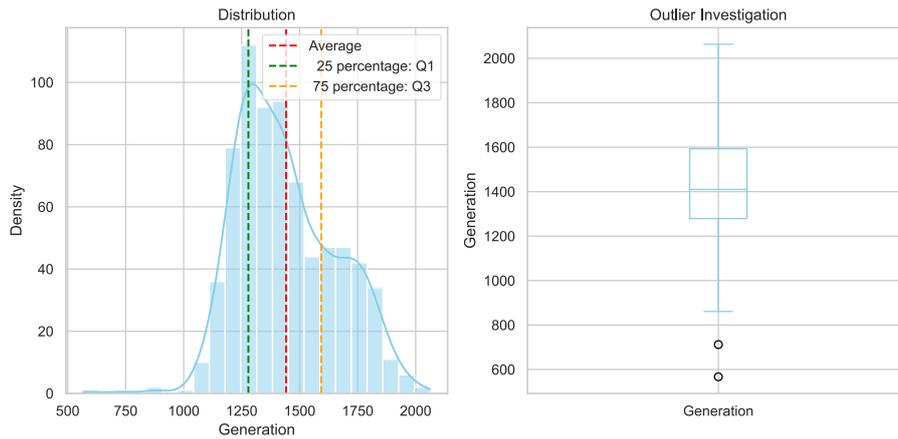

Fig.3. Generation distribution and outlier investigation

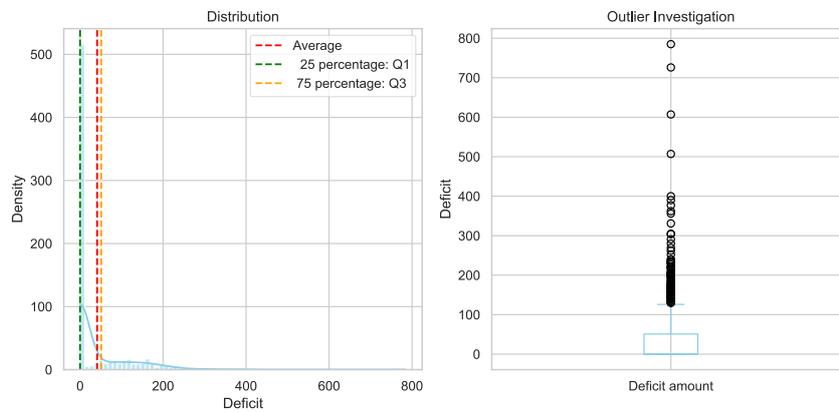

Fig.4. Deficit distribution and outlier investigation

The box plot highlights extreme outliers exceeding 700 kW, likely linked to imbalances between load and generation. Outliers in the generation data, where generation falls below 800 kW, coincide with periods of substantial deficits. These shortfalls, possibly due to planned outages or unforeseen events, create a gap between supply and demand, particularly during peak load times, emphasizing the system's vulnerability to significant shortfalls when generation capacity is insufficient.

### C. Heatmap of Load, Generation, and Deficit

To visualize temporal trends and potential relationships between electricity load, generation, and deficit, we analyzed the two-year heatmaps for each variable.

- *Electricity Loads:* As shown in Fig. 5, electricity demand exhibits notable seasonal variations. August sees a significant surge in demand, likely due to increased cooling needs, while January and February experience lower demand. In 2023, the heatmap reveals an overall increase in electricity demand compared to 2019, with a smoother weekly pattern. The contrast between weekday and weekend consumption observed in 2019 is less pronounced in 2023, indicating a shift toward more uniform demand, possibly due to changing residential or industrial consumption patterns. The more stable load pattern in 2023 could reflect improved grid infrastructure or shifts in energy usage behavior.

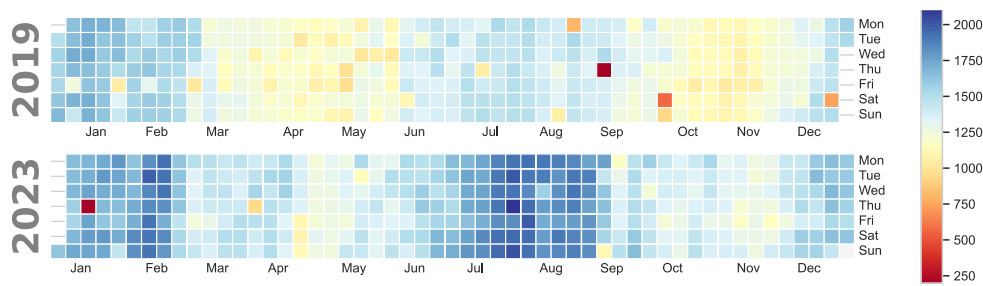

Fig.5. Heatmap of load patterns for 2019 and 2023

Data-driven Insights for Informed Decision-Making: Applying LSTM Networks for Robust Electricity Forecasting in Libya- *Electricity Generation:* As shown in Fig. 6, generation in 2019 fluctuated significantly, with periods of both high and low output. Notably, August sees higher generation levels to meet peak demand. However, dips below 1200 kW, particularly in the first quarter, correlate with the fluctuations in load, contributing to supply-demand imbalances. In contrast, 2023 presents a more stable generation output, ranging between 1200 kW and 1400 kW, with fewer drastic fluctuations. Despite higher and more consistent generation, 2023 shows less alignment between peak generation and peak load, indicating a shift in dynamics between generation and demand.

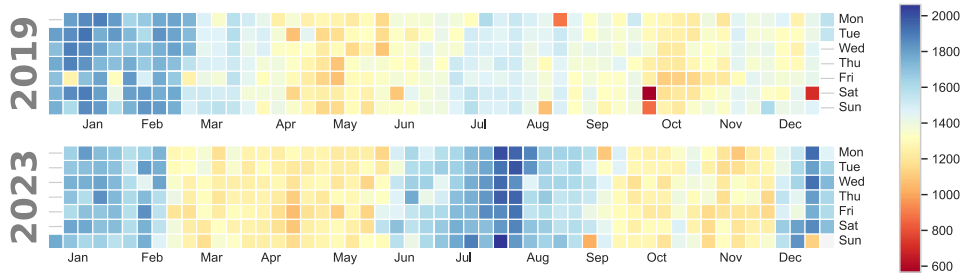

Fig.6. Heatmap of generation patterns for 2019 and 2023

- *Electricity Deficit:* In 2019, as shown in Fig. 7, significant and frequent deficits exceeding 500 kW were observed, especially in high-demand periods like August. These deficits align with the generation dips below 1200 kW, suggesting that inconsistent generation struggled to meet fluctuating demand. However, in 2023, the deficit heatmap shows a marked improvement, with minimal deficits and fewer occurrences exceeding 100 kW. This reduction in deficits corresponds with the more stable and consistent generation and load patterns, highlighting the system's improved capacity to meet demand.

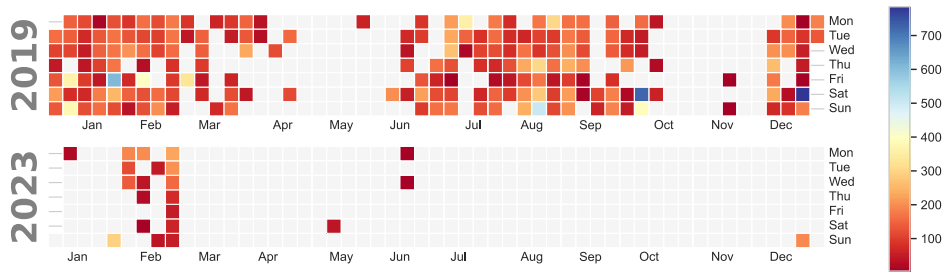

Fig.7. Heatmap of deficit patterns for 2019 and 2023

The stark contrast between the 2019 and 2023 heatmaps underscores the importance of enhanced generation capacity and a more predictable load profile in reducing deficits and ensuring a stable electricity supply.

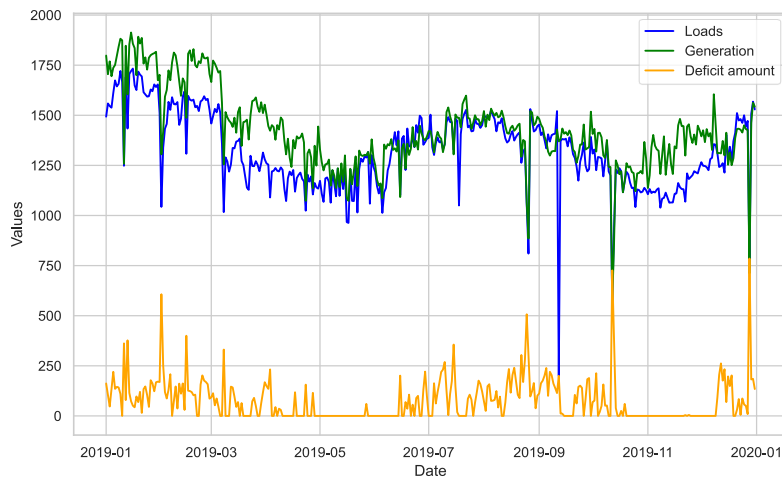

Fig.8. The trend analysis of load, deficit, and generation during 2019

## D. Trend Analysis of Load, Generation, and Deficit

As shown in Fig. 8, the 2019 electricity data reveal a highly unstable system, with significant fluctuations in both load (demand) and generation. While generation attempts to match demand, mismatches often lead to substantial deficits,

Data-driven Insights for Informed Decision-Making: Applying LSTM Networks for Robust Electricity Forecasting in Libya

especially during peak load periods exceeding 1750 kW. Notably, dips in generation, particularly in the first quarter and September, cause deficits exceeding 500 kW. These imbalances suggest challenges in maintaining a stable equilibrium between supply and demand, likely driven by seasonal variations and potential generation capacity limitations.

In contrast, Fig. 9, for 2023 shows a more stable electricity system. While both load and generation still fluctuate, generation more closely aligns with demand throughout the year. The deficit remains minimal for most of the year, with only a few minor spikes. This indicates improvements in generation capacity and more predictable consumption patterns, leading to a more reliable and balanced electricity system in 2023.

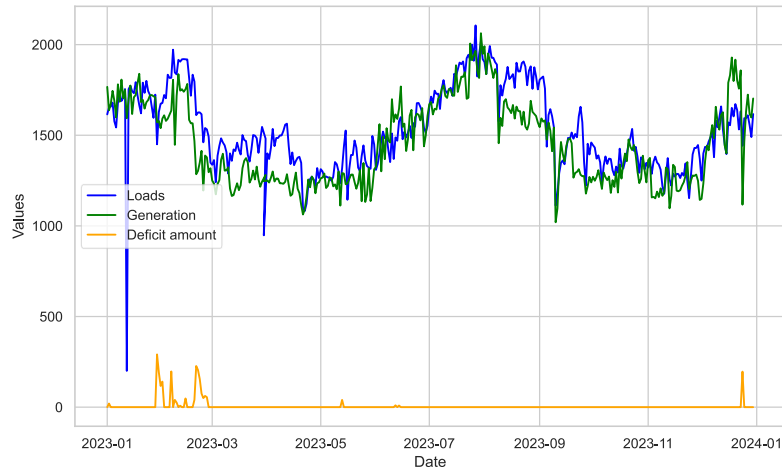

Fig.9. The trend analysis of load, deficit, and generation during 2023

E. *Seasonality Analysis of Load, Generation, and Deficit*

- *Time Series Stationarity of Loads*: The seasonal analysis of load time series for 2019 and 2023 shown in Fig. 10 shows clear non-stationarity, with downward and upward trends, respectively. The ADF test confirms this, with p-values of 0.129 for 2019 and 0.127914 for 2023, exceeding the 0.05 threshold. This suggests non-stationarity in both years, further supported by slowly decaying ACF plots. These results emphasize the need for transformations, like differencing or seasonal decomposition, to stabilize the mean and variance before forecasting.
- *Post-Differencing Stationarity of Loads*: After applying first-order differencing, Fig. 11 demonstrates that the electricity generation data is converted into a stationary series. The time series plot reveals that the differenced data oscillates around a mean of zero. Additionally, the ACF and PACF plots exhibit a quick decay and diminished reliance on past values, signifying that the series has been effectively stabilized and is now appropriate for forecasting.

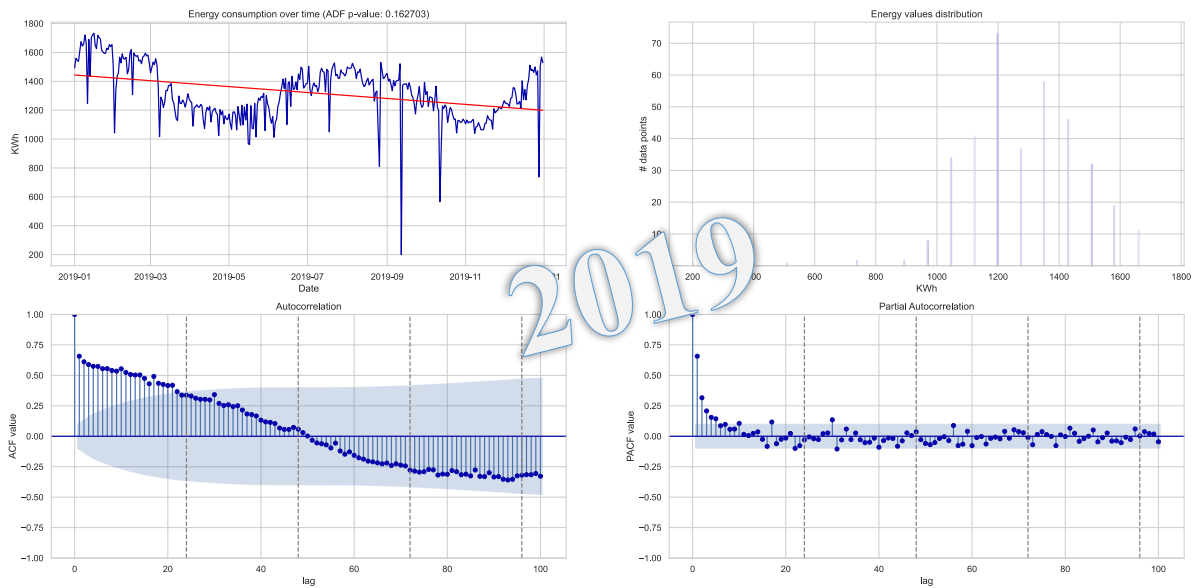

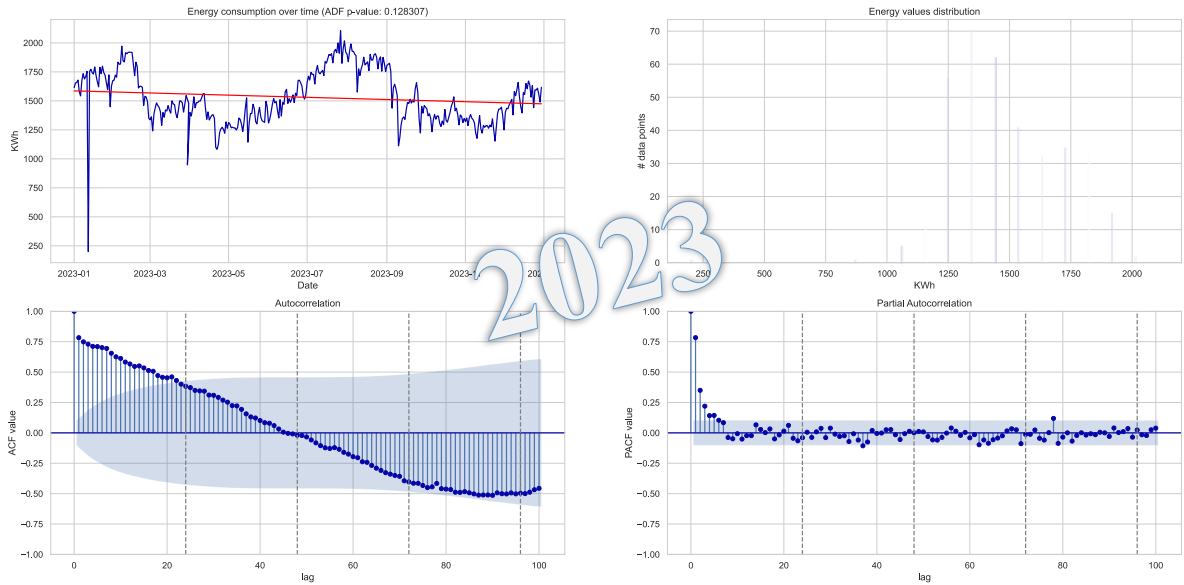

Fig.10. Load distribution, dynamic trend, and seasonal patterns 2019-2023

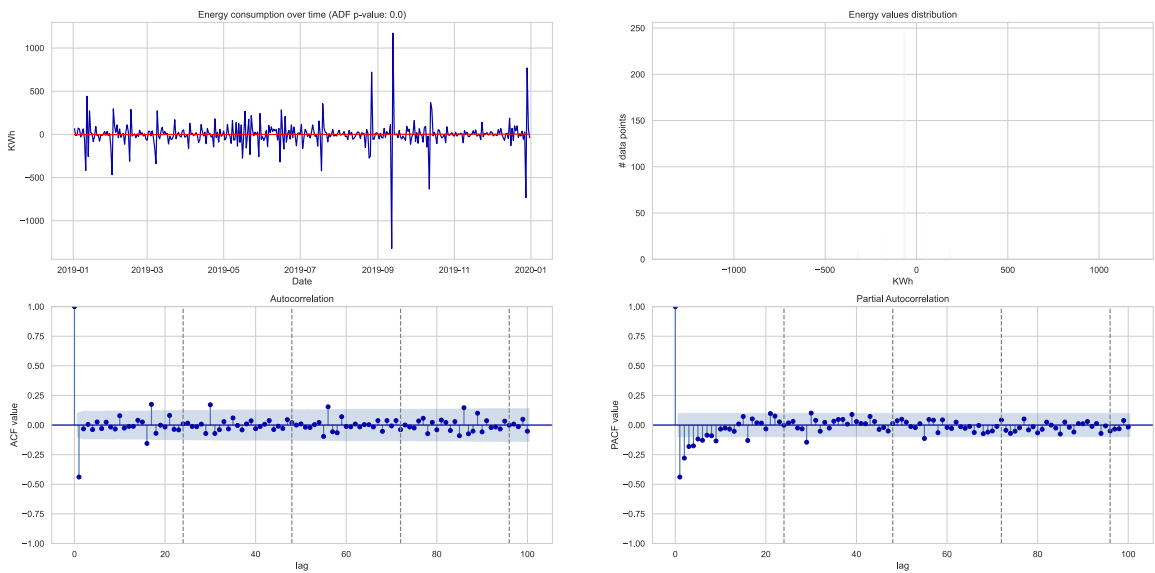

Fig.11. Confirmation of stationarity in electricity load data after differencing

- Time Series Stationarity of Generation: As shown in Fig. 12, the 2019 and 2023 electricity generation data exhibit potential non-stationarity, with a visible upward trend. The ADF test for 2023 yields a p-value of 0.204755, confirming non-stationarity. The ACF plot also shows significant autocorrelation, reinforcing this conclusion. First-order differencing or seasonal decomposition is recommended to stabilize the data for forecasting models.
- *Post-Differencing Stationarity of Generation:* Following first-order differencing, Fig. 13 confirms the transformation of electricity generation into a stationary series. The differenced data fluctuates around a mean of zero, as seen in the time series plot. The ACF and PACF plots show rapid decay and reduced dependence on past values, indicating successful stabilization of the series, making it suitable for forecasting.
- *Time Series Stationarity of Deficit:* The 2019 and 2023 electricity deficit data Fig. 14 also show non-stationary characteristics. The ADF test for 2019 yields a p-value of 0.0902, indicating non-stationarity. ACF plots further support this, with slow decay and significant autocorrelation. This suggests that differencing or seasonal decomposition is necessary before applying forecasting models.
- *Post-Differencing Stationarity of Deficit:* After applying first-order differencing, as shown in Fig. 15, the electricity deficit data becomes stationary. The time series plot shows fluctuations around zero, and the ADF test yields a p-value of 1.55e-13, rejecting the null hypothesis of non-stationarity. The ACF and PACF plots confirm this, demonstrating rapid decay and reduced dependence on past values, indicating the data is now suitable for forecasting.

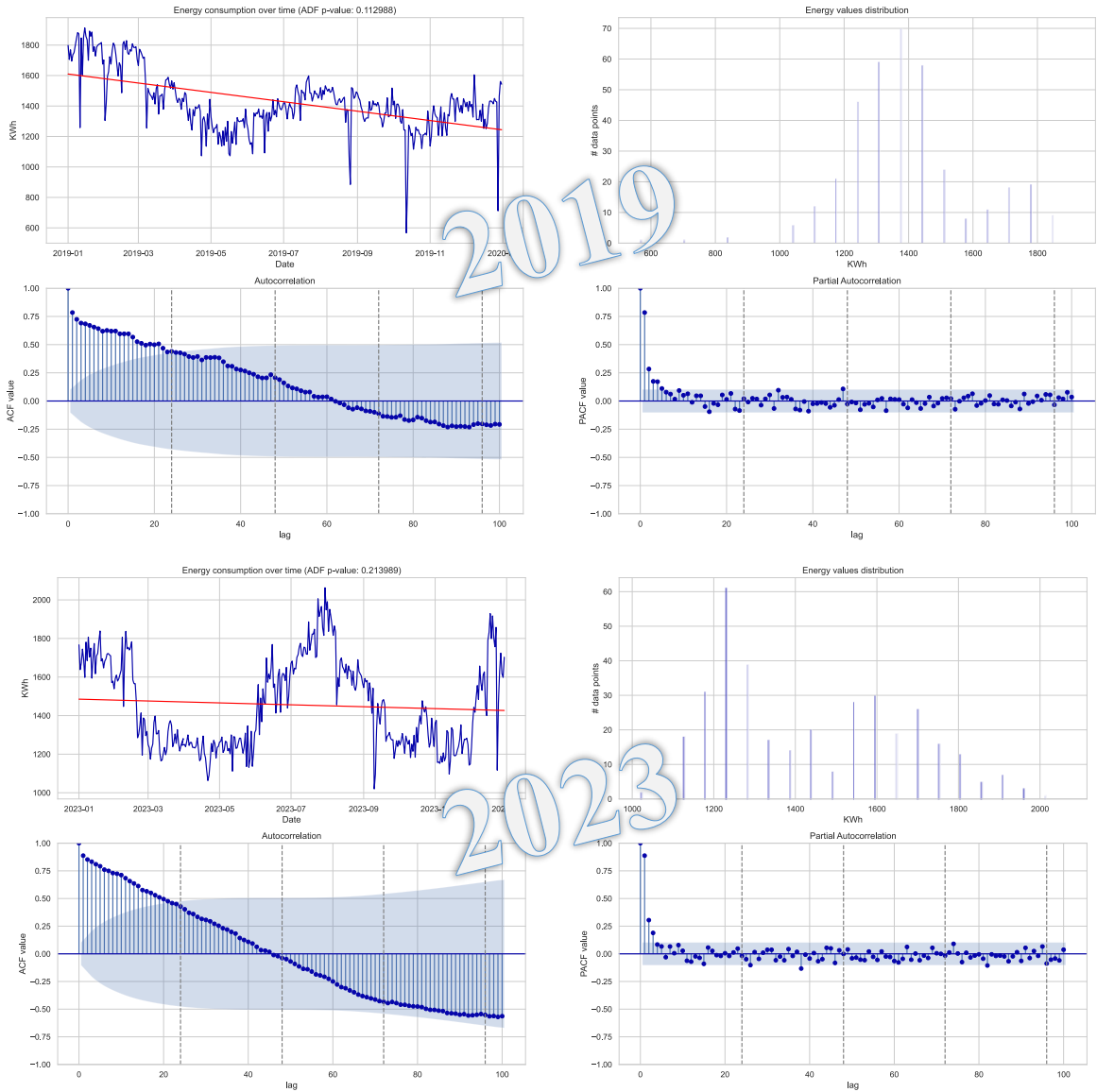

Fig.12. Generation distribution, dynamic trend, and seasonal patterns during 2019-2023

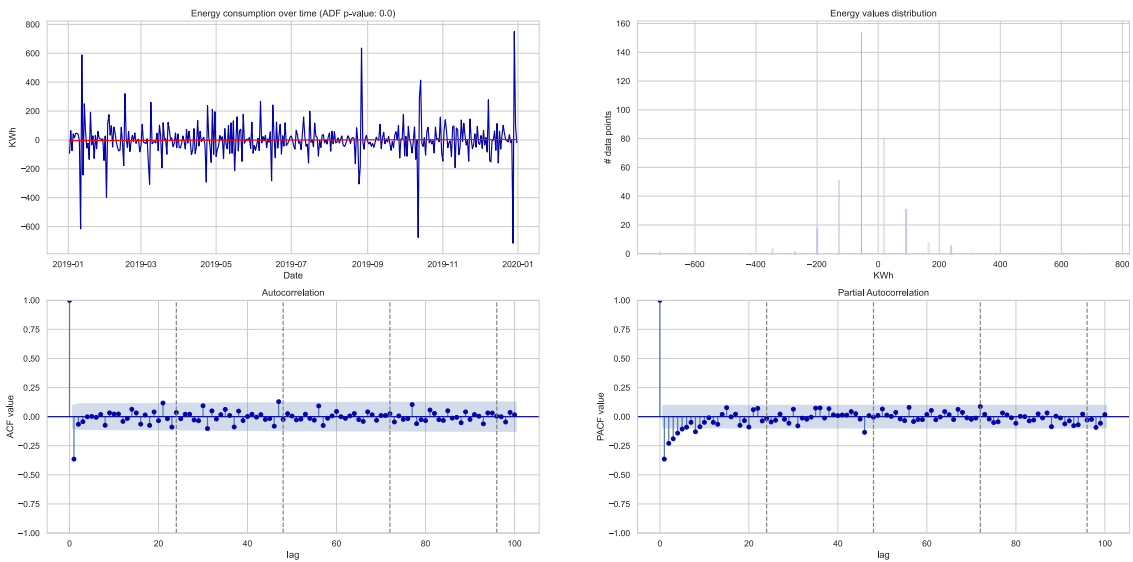

Fig.13. Confirmation of stationarity in electricity generation data after differencing

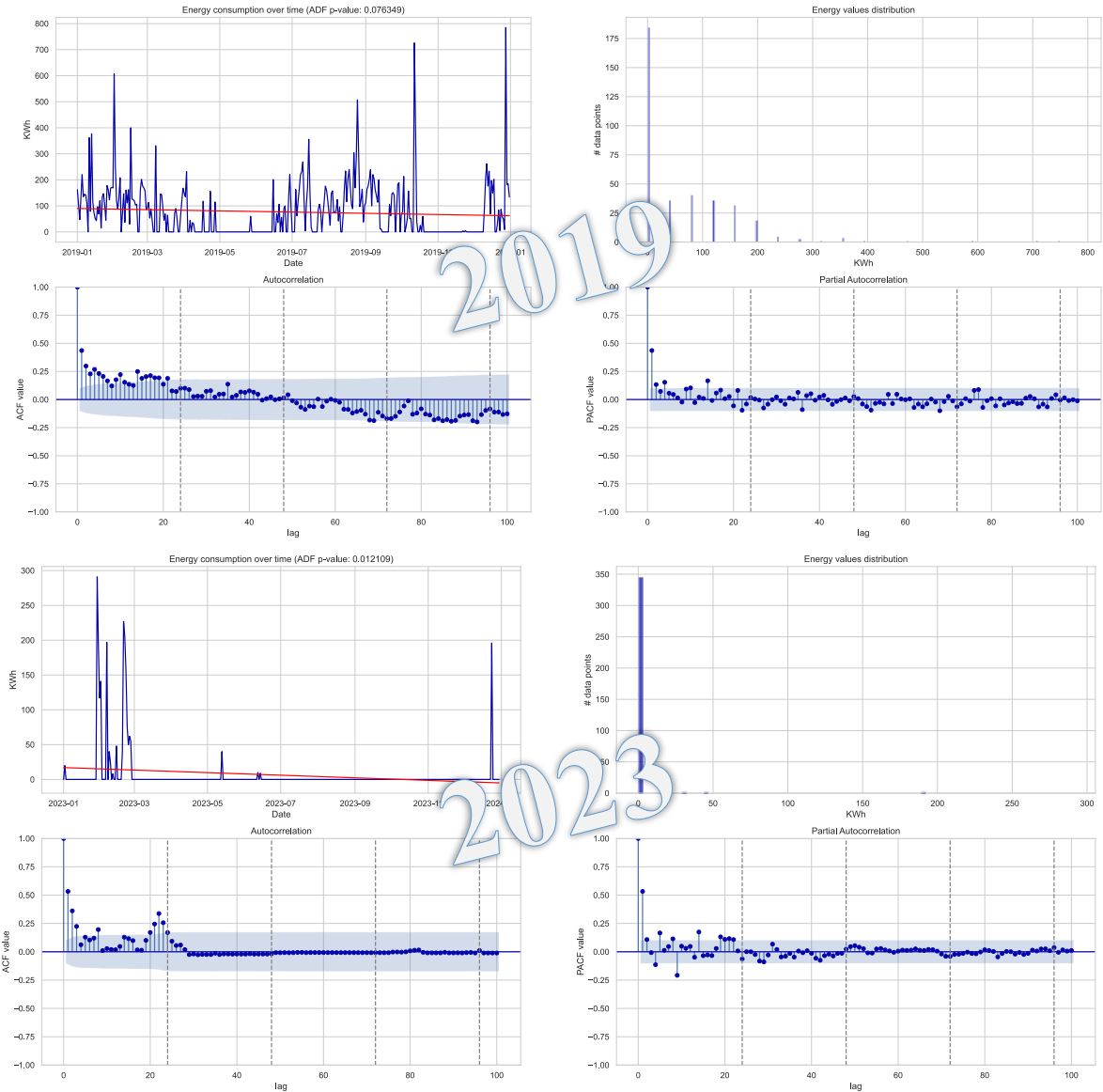

Fig.14. Deficit distribution, dynamic trend, and seasonal patterns 2019-2023

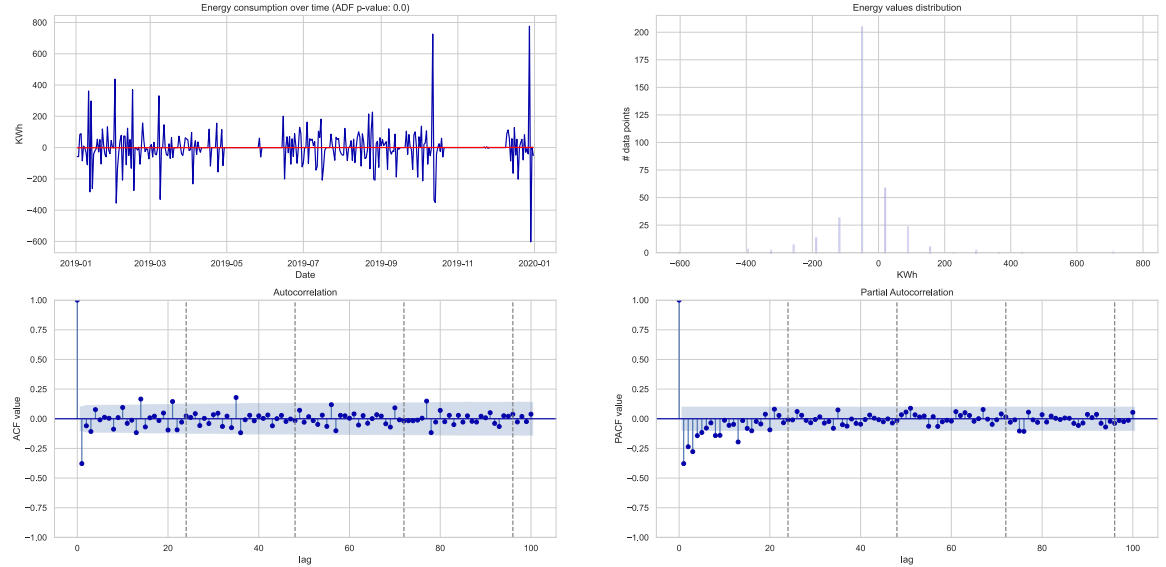

Fig.15. Confirmation of stationarity in electricity loads data after differencing

Data-driven Insights for Informed Decision-Making: Applying LSTM Networks for Robust Electricity Forecasting in Libya

The analysis of electricity data from 2019 and 2023 shows improved system stability in 2023, with more consistent generation aligning better with demand, reducing deficits. First-order differencing transformed non-stationary time series data into stationary series, enhancing forecasting reliability. These improvements in generation infrastructure and consumption behavior contributed to a more stable electricity system and more accurate predictions.

*F. Fitting the Models*

Our objective to create accurate electricity forecasting models led us through an extensive hyper-parameter optimization process for ARIMA, SARIMA, ARIMAX, LSTM, and XGBoost models, focusing on load, deficit, and generation. The following sections outline the findings:

- Fitting the ARIMA Models

We aimed to identify optimal ARIMA model configurations that balanced model fit and complexity, using AIC and BIC as key indicators. The results summarized in Table 3, highlighting several key insights:

- Load Forecasting: The ARIMA (2,1,2) model emerged as the best for load forecasting, achieving the lowest AIC (5139) and a competitive BIC (5162), suggesting that both autoregressive and moving average components, along with first-order differencing, effectively captured electricity load dynamics.
- Deficit Forecasting: The SARIMA (3,0,0) model performed best for deficit forecasting, with the lowest AIC (6277). Despite a slightly lower sigma2 for the ARIMA model, its higher AIC and BIC indicated overfitting, making SARIMA the more suitable choice.
- Generation Forecasting: For electricity generation, the ARIMA (1,1,1) model achieved the lowest AIC (8754) and a comparable BIC (8768), suggesting its efficacy in capturing generation patterns. The SARIMA model did not offer significant improvement.

Model Selection: Based on these findings, we selected the ARIMA (2,1,2), SARIMA (3,0,0), and ARIMA (1,1,1) models as the leading choices for predicting load, deficit, and generation, respectively.

Table 3. ARIMA based models parameters setting

| Model | Parameter | Load | Deficit | Generation |
|---|---|---|---|---|
| ARIMA | (p, d, q) | (2, 1, 2) | (2, 0, 1) | (1, 1, 1) |
| | AIC | 5139 | 8238 | 8754 |
| | BIC | 5162 | 8261 | 8768 |
| | Sigma2 | 1.60e+04 | 4669.9443 | 1.02e+04 |
| SARIMA | (p, d, q) | (3, 0, 1) | (3, 0, 0) | (0, 1, 2) |
| | Seasonal Order | (1, 2, 0, 12) | (0, 0, 1, 12) | (1, 1, 0, 12) |
| | AIC | 9144 | 6240 | 8957 |
| | BIC | 9140 | 6277 | 8956 |
| | Sigma2 | 1.51e+04 | 4511.618 | 1.093e+04 |
| ARIMAX | (p, d, q) | (1, 1, 0) | (1, 0, 2) | (2, 1, 0) |
| | Exogenous Variables | Temperature, Humidity | Temperature, Humidity | Temperature, Humidity |
| | AIC | 9402 | 7244 | 8822 |
| | BIC | 9139 | 7200 | 8857 |
| | Sigma2 | 1.66e+04 | 4511.808 | 1.14e+04 |

- Fitting the XGBoost Model

We optimized the XGBoost model through a grid search over key hyper-parameters: number of estimators, learning rate, and tree max depth, the search explored various options, and the optimal configuration emerged as shown in Table 4.

Table 4. XGBoost hyper-parameter optimization

| Hyper-parameter | Options Tested | Optimal Value |
|---|---|---|
| Number of estimators | 100, 500 | 200 |
| Learning rate | 0.01, 0.1 | 0.01 |
| Tree max depth | 3, 5, 7 | 3 |

This configuration, achieving the lowest negative mean squared error (21.076) during 5-fold cross-validation, will be evaluated on the test set for predictive accuracy on future electricity data.

- Fitting the LSTM Model

The LSTM model optimized by conducting a grid search to determine the optimal number of memory cells (units). We tested three options (50, 100, and 200 units) to balance complexity and accuracy. The grid search revealed that 100 units provided the best performance. The final architecture includes 100 LSTM units, ReLU activation for non-linearity, and the Adam optimizer for efficient training.

- XGBoost and LSTM Training and Testing

To prepare for forecasting, the historical data split into training and testing sets. The training data, covering January 1, 2019, to April 30, 2023, provided a substantial period for the models to capture underlying patterns. The test data, from May 1, 2023, to December 30, 2023, used to evaluate the predictive accuracy of the trained models on unseen data.

G. *Performance Comparison of Models*

- Performance of Models for Loads Prediction:

From Table 5, it is evident that the LSTM model significantly outperforms all other models across all evaluation metrics. With exceptionally low MSE, RMSE, MAE, and MAPE, the LSTM demonstrates a superior ability to capture the complexities of electricity load patterns. The ARIMA and XGBoost models show similar performance, with XGBoost slightly outperforming ARIMA in MSE and RMSE, but based on MAPE is opposite. Both models outperform the Dynamic ARIMA, SES, and SARIMA models, showing a clear advantage in accuracy.

Table 5. Models performance for forecasting the load demand

| Model | MSE | RMSE | MAE | MAPE (%) |
|---|---|---|---|---|
| LSTM | 0.018212 | 0.134951 | 0.103151 | 0.006267 |
| ARIMA | 78954.8816 | 280.9891 | 210.3705 | 12.5202 |
| XGBoost | 74732.6432 | 273.3727 | 212.3116 | 12.7191 |
| Dynamic ARIMA | 101963.597 | 319.31739 | 240.3206 | 14.2866 |
| ESE | 114922.741 | 339.0025 | 266.1475 | 15.9188 |
| SARIMA | 304647.0960 | 551.9484 | 497.9347 | 31.3871 |

The visualization in Fig. 16 confirms these findings. The LSTM predictions closely align with the actual load values. While ARIMA and XGBoost forecasts show reasonable fit, they are less accurate than LSTM. In contrast, the Dynamic ARIMA, SES, and especially the SARIMA models exhibit significant deviations from the actual load patterns, highlighting their lower predictive accuracy.

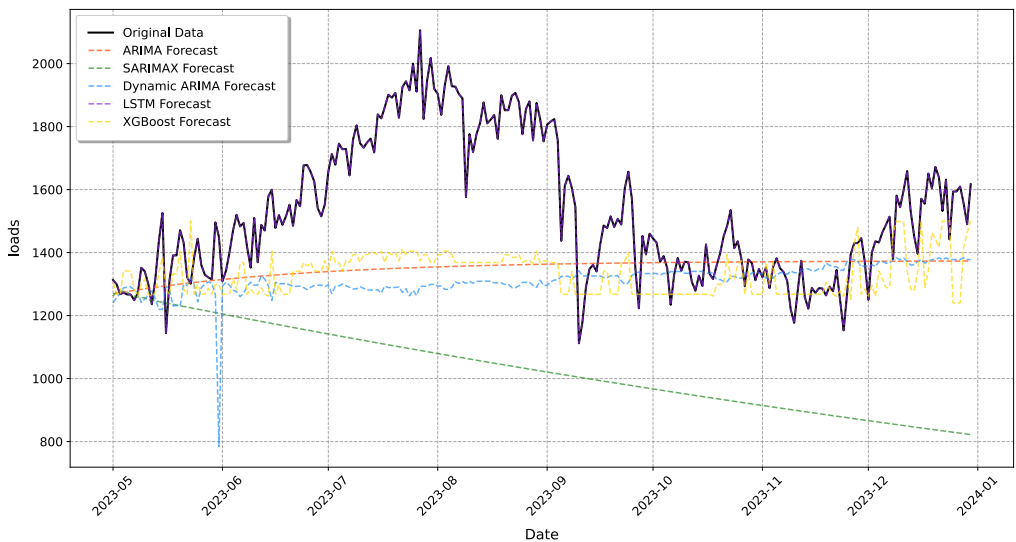

Fig.16. Comparing model performance in forecasting load

The LSTM model stands out as the best performer in the electrical load-forecasting task, capturing the overall trends and fluctuations, especially during high-volatility periods like August and December. XGBoost also performs well, handling the load patterns and spikes, particularly in August, though it falls slightly behind LSTM in terms of accuracy. ARIMA delivers decent results, effectively capturing the overall trend, but tends to underestimate the load during peak periods, making it less accurate than LSTM and XGBoost. SARIMA, on the other hand, is the weakest model, consistently underestimating the load with a noticeable downward bias, likely due to ineffective use of exogenous variables such as temperature and humidity. Dynamic ARIMA suffers from instability, often overreacting to data fluctuations and producing forecasts far from the actual values, particularly during volatile periods. Finally, SES proves too simplistic for this dataset, failing to capture the complex patterns in the load and resulting in a flat and inaccurate forecast.

- Performance of Models for Generation Prediction:

As shown in Table 6, the LSTM model excels in forecasting electricity generation, outperforming all other models across the four-evaluation metrics with remarkably low MSE, RMSE, MAE, and MAPE values. This highlights its superior ability to capture the complex patterns in electricity generation. XGBoost and ARIMA show similar performance, with XGBoost slightly outperforming ARIMA in most metrics, except MAPE. Both models perform considerably better than the Dynamic ARIMA and SARIMA models, which show significant deviations from actual generation values. The visualization in Figure 16 further confirms these findings, with LSTM forecasts closely matching actual generation, while XGBoost and ARIMA predictions show a larger deviation, and Dynamic ARIMA and SARIMA models diverge even further, indicating poorer accuracy.

Table 6. Models performance for forecasting the generation

| Model | MSE | RMSE | MAE | MAPE (%) |
|---|---|---|---|---|
| LSTM | 0.028676 | 0.169339 | 0.145829 | 0.01001 |
| XGBoost | 61028.93739 | 247.0403 | 193.2630 | 12.1832 |
| ARIMA | 61172.30146 | 247.3303 | 201.0460 | 13.1059 |
| Dynamic ARIMA | 75069.63355 | 273.9883 | 225.1083 | 15.3582 |
| SARIMA | 256615.127 | 506.571 | 443.7787 | 28.6439 |

In terms of strengths, LSTM stands out for its ability to capture the overall generation pattern, including downward trends and fluctuations, particularly during periods of high volatility. XGBoost performs well too, capturing major fluctuations but slightly overestimating the generation towards the end of the forecast horizon. On the other hand, SARIMA consistently shows a downward trend, deviating significantly from actual generation, suggesting misinterpretation of the data. ARIMA also exhibits a downward trend, but with slightly better alignment with actual generation compared to SARIMA. Dynamic ARIMA struggles to provide stable forecasts, overreacting to data fluctuations and producing predictions far from actual values.

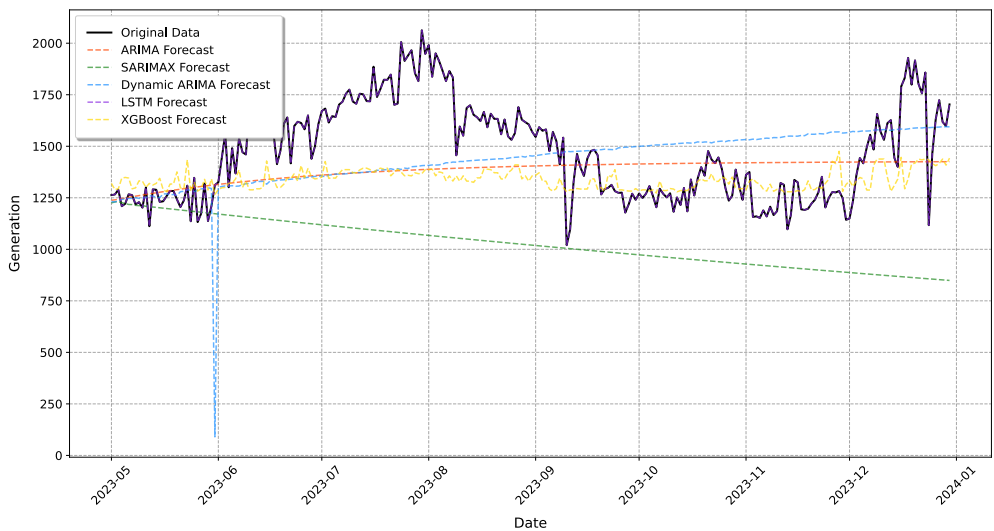

Fig.17. Comparing model performance in forecasting generation

- Performance of Models for Deficit Prediction:

As presented in Table 7, the LSTM model outperforms all other models in forecasting electricity deficit, achieving substantially lower MSE, RMSE, MAE, and MAPE values, indicating its ability to capture the complex, non-linear patterns of the deficit data. In contrast, other models exhibit much higher error metrics, showing their limitations in predicting deficit values accurately. Although the SARIMAX model shows lower error than XGBoost, Dynamic ARIMA, and ARIMA, it still falls behind the LSTM in terms of performance. The visualization in Fig. 18 further supports these conclusions, with LSTM forecasts closely aligning with the actual deficit values, even capturing the significant spike in December. Other models, particularly Dynamic ARIMA and ARIMA, struggle to capture this spike and deviate significantly from the actual values.

Table 7. Models performance for forecasting the deficit

| Model | MSE | RMSE | MAE | MAPE (%) |
|---|---|---|---|---|
| LSTM | 0.086256 | 0.29369 | 0.28686 | 5.17726 |
| SARIMAX | 164.7418 | 12.8351 | 1.04508 | 100.000 |
| XGBoost | 3805.091 | 61.6854 | 53.2342 | 124.617 |
| Dynamic ARIMA | 5295.786 | 72.7721 | 48.8081 | 220.371 |
| ARIMA | 3151.352 | 56.1369 | 54.8220 | 228.013 |

The challenge in forecasting electricity deficit is largely due to the inherent volatility and the infrequent occurrence of large deficits, which makes it difficult for traditional time series models and tree-based models like XGBoost to capture the underlying patterns.

In terms of strengths, LSTM excels in forecasting the major spikes in the deficit data (such as in June and December), effectively handling non-linear relationships and learning from infrequent but significant events. XGBoost performs reasonably well, maintaining a flat forecast close to zero, although it misses the large spikes. This may still be useful for identifying significant deficit periods. In contrast, ARIMA, SARIMA, and Dynamic ARIMA models fail to capture the significant spikes, consistently predicting deficits near zero, which makes them ineffective in identifying periods of high deficit and highlights the limitations of traditional models in forecasting extreme, infrequent events.

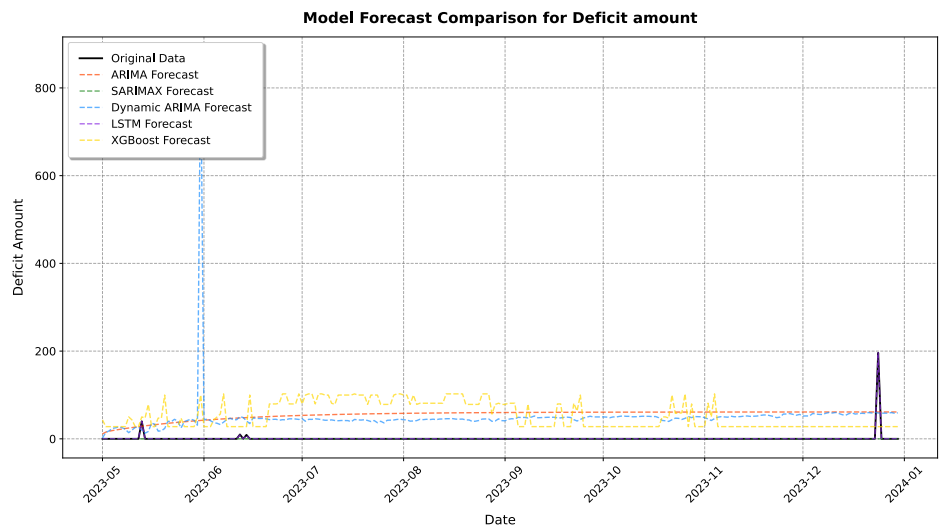

Fig.18. Comparing model performance in forecasting deficit

- Summary of the results obtained

The LSTM model consistently outperformed traditional statistical models (ARIMA, SARIMA, ARIMAX) and machine learning techniques (XGBoost, SES) across the three forecasting tasks: load, deficit, and generation. For electricity load forecasting, the LSTM achieved a remarkably low MAPE of 0.006%, significantly outperforming ARIMA (12.52%) and XGBoost (12.72%). The model demonstrated a strong ability to capture complex non-linear patterns and long-term dependencies, closely aligning with actual load values, even during high volatility periods. In electricity generation forecasting, LSTM achieved a MAPE of 0.01%, compared to ARIMA's 13.11% and XGBoost's 12.18%. The model accurately predicted generation trends, including seasonal fluctuations and extreme events, demonstrating its robustness in handling non-stationary data. For electricity deficit forecasting, LSTM achieved a MAPE of 5.18%, outperforming SARIMAX (100%) and XGBoost (124.62%). Despite the inherent volatility and infrequent large deficits, LSTM effectively captured extreme events, such as the significant spike in December.

The superior performance of the LSTM model aligns with findings from recent studies that highlight the effectiveness of deep learning techniques in electricity forecasting. For load forecasting, studies such as [24] and [25] have demonstrated that LSTM outperforms traditional models by capturing short- and long-term dependencies in electricity load data. In generation forecasting, research by [26], [27], and [28] has shown that LSTM excels at predicting renewable energy generation, particularly in scenarios with high volatility and non-linear patterns. While deficit forecasting is less studied, this study's results suggest that LSTM can handle the challenges posed by infrequent but significant deficit events. This research also goes beyond existing studies by tailoring the LSTM model to the unique electricity patterns of NBPP, incorporating exogenous variables (temperature, humidity) to enhance forecasting accuracy, and providing actionable insights for improving grid reliability and energy management in regions facing similar challenges.

*H. Future Forecasting*

Following a thorough evaluation of various forecasting models for electricity load, generation, and deficit, the LSTM) neural network consistently outperformed all other models. Across all three-target variables, the LSTM demonstrated exceptionally low error metrics, reflecting its high accuracy in tracking actual values, as seen in the corresponding visualizations. This performance underscores LSTM's superior capability to capture the complex, non-linear patterns and intricate temporal dependencies inherent in electricity data, which traditional methods struggle to handle. Notably, the LSTM model also excelled in managing volatility and infrequent events, which are often present in energy datasets. Given its robust performance, we have chosen the optimized LSTM model for future forecasting tasks. This model employed to predict electricity load, deficit, and generation for the years 2024 until 2025, providing critical insights to support informed decision-making and strategic planning in future energy management.

- LSTM Load Forecasting in 2025

As shown in Fig. 19. The LSTM's predictions for electricity load for the 2025 (depicted as the red line) compared against the historical load data (blue line), which extrapolated based on the average values observed during the model's training period. The LSTM model successfully captures the cyclical nature of the data, predicting a sinusoidal wave for each year, with peaks typically occurring mid-year and troughs at the beginning and end. This accuracy in modeling seasonality highlights the model's ability to learn from historical data trends and project them into future periods. However, the LSTM's predictions are notably smoother than the historical data, reflecting the model's focus on long-term trends and seasonal cycles while downplaying short-term fluctuations. This characteristic is expected, as forecasting models often prioritize capturing dominant patterns to avoid overfitting to noise or random variations present in the historical dataset. This approach ensures the model's robustness in providing generalizable and reliable forecasts for strategic energy planning.

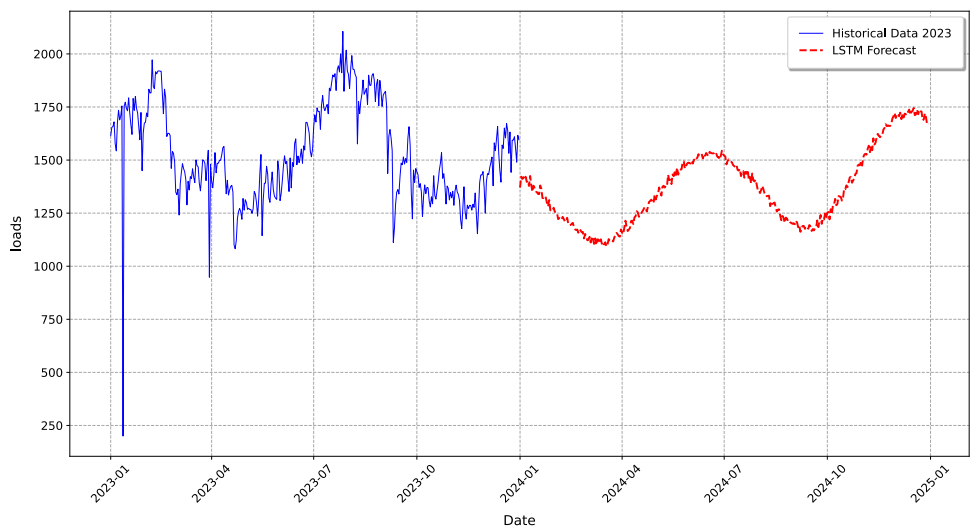

Fig.19. Load forecasting for the 2025

- LSTM Generation Forecast in 2025:

Fig. 20 illustrates the LSTM model's predictions for electricity generation from 2024 to 2025 (red line), compared to the extrapolated historical generation data (blue line), based on the average values observed during the training period. The LSTM model successfully identifies the cyclical patterns inherent in the historical data, forecasting a sinusoidal wave for each year, with peaks occurring around mid-year and troughs at the beginning and end. This effective representation of seasonality reflects the model's ability to learn from historical generation trends and project them into the future. Similar to the load forecast, the LSTM's generation predictions exhibit a noticeable smoothing effect, prioritizing the general trend and seasonal fluctuations while downplaying finer, short-term variations in the historical data. This behavior indicates that the model focused on capturing the dominant, recurring patterns in electricity generation, which helps reduce the impact of potentially noisy fluctuations, resulting in a smoother, more stable forecast curve. The model's ability to capture these overarching trends provides valuable insights for future energy planning and generation management.

- LSTM Deficit Forecast in 2025:

As previously shown the LSTM model perform well in forecasting electricity loads and generation, while highlighting its inability to accurately predict the electricity deficit for the period from 2024 to 2025 as shown in Fig. 21. The model demonstrates strong predictive capabilities for loads and generation, capturing their patterns effectively and providing valuable insights. However, when applied to deficit forecasting, it struggles to provide accurate predictions, compared to the historical deficit data.

Data-driven Insights for Informed Decision-Making: Applying LSTM Networks for Robust Electricity Forecasting in Libya

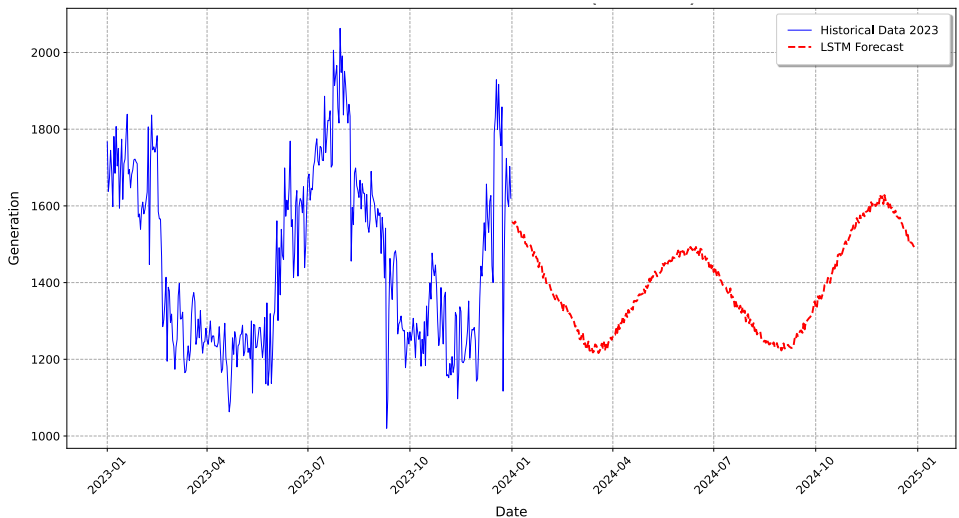

Fig.20. Generation forecasting for the 2025

The challenges in deficit forecasting compounded by the presence of numerous outliers in the data and the irregular occurrence of deficits. Despite trained on historical deficit data, the model exhibits limitations in capturing short-term volatility and erratic fluctuations. The cyclical pattern observed in the historical data, which reflects underlying seasonality, which not adequately mirrored in the model's deficit predictions.

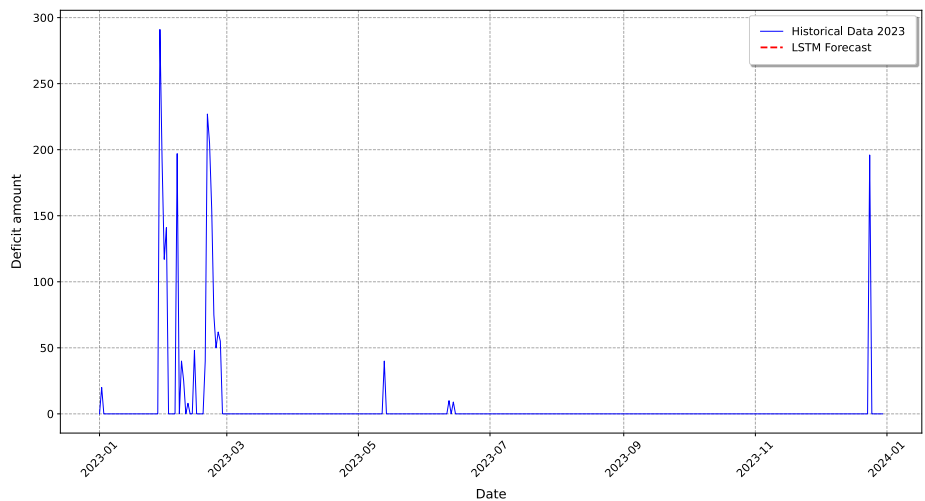

Fig.21. Deficit forecasting for the 2025

## 5. Conclusions

This study addresses the critical challenge of electricity forecasting in NBPP in Benghazi, Libya, a region plagued by frequent load shedding, generation deficits, and aging grid infrastructure. By leveraging advanced machine learning techniques, particularly Long Short-Term Memory (LSTM) networks, the study develops a robust and accurate forecasting framework tailored to the unique electricity consumption and generation patterns of the region. The LSTM model consistently outperformed traditional statistical models (ARIMA, SARIMA, ARIMAX) and machine learning techniques (XGBoost, SES) across all forecasting tasks—load, deficit, and generation—demonstrating its ability to capture complex non-linear patterns, long-term temporal dependencies, and inherent volatility. With remarkably low MAPE values of 0.006%, 0.01%, and 5.17% for load, deficit, and generation forecasting, respectively, the LSTM model effectively predicted electricity trends. This performance provides actionable insights for energy policymakers and grid operators, enabling proactive resource allocation, improved demand-side management, and enhanced grid resilience. The integration of exogenous variables, such as temperature and humidity, further enhanced forecasting accuracy, offering a more comprehensive understanding of the factors influencing electricity demand and supply. This research highlights the potential of advanced machine learning techniques to address energy forecasting challenges, providing a pathway for more reliable and sustainable electricity management, particularly in regions with similar supply constraints.

## 6. Recommendations

Based on the findings of this study, several key recommendations proposed to enhance energy management and grid operations in Benghazi and beyond. First, grid operators and energy policymakers should adopt advanced forecasting models like LSTM to improve the accuracy of load, deficit, and generation predictions. Additionally, incorporating exogenous variables such as temperature and humidity into forecasting models should be prioritized to improve forecasting accuracy and provide more actionable insights. Second, investments in advanced metering infrastructure and data acquisition systems are crucial to enable real-time data collection, improve model responsiveness, and optimize forecasting outcomes. High-quality historical data is essential for training and refining machine learning models to ensure their accuracy and reliability. Third, accurate load and generation forecasts should be used to optimize resource allocation, reducing the risk of supply shortages or over-generation. Demand-side management strategies, such as load shifting or incentive programs, should be implemented to balance supply and demand during peak periods. Finally, future research should explore integrating additional exogenous variables, such as economic indicators or population growth, to further enhance forecasting accuracy. The development of hybrid models combining the strengths of LSTM and other machine learning techniques could address specific forecasting challenges and improve overall model performance. Furthermore, applying the proposed framework to other regions with similar energy supply challenges should be investigated to validate its scalability and generalizability.

## Authors' Profiles

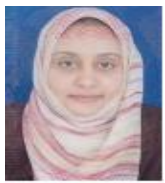

**ASMA AGAAL** is an Assistant lecturer at the Faculty of Technical Sciences, Sabha, Libya. She holds a Master's degree in Computer Science (Faculty of Science, University of Sebha in 2023). Her research interests include pattern recognition, image processing, machine learning methods, deep learning, and time series analysis. Published multiple research papers in the fields of artificial intelligence, computer science, and related disciplines. Actively participated in various local and international conferences and seminars, showcasing expertise and advancements in artificial intelligence.

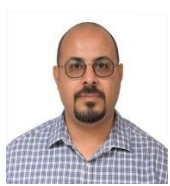

**Dr. Mansour Essgaer** earned his Ph.D. in Computer Science from the National University of Malaysia, in 2015. Since 2016, he has been serving as an Assistant Professor in the Faculty of Information Technology at Sebha University, Libya. His research interests encompass data mining, artificial intelligence, machine learning, natural language processing, and combinatorial optimization. He is an active member of the IEEE and has contributed to numerous publications in his field.

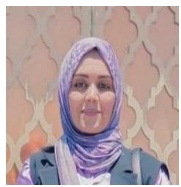

**Hend Farkash** Assistant Professor, Lecturer in The College OF Electrical & Electronic Technology, Benghazi, Libya, and a collaborator in several private universities. Education and Qualification: have Master degree in Computer Sciences. Specialization: Artificial Intelligence, Graduation Year :2010, University: faculty of information technology - Benghazi University, and have B.Sc. in computer science. Graduation Year :2003 University: faculty of computer science - Benghazi University, Conferences and Seminars: have several researches in the field of intelligence, computer science and artificial intelligence, and distinguished participation in local and international conferences, In addition, and a member of the IEEE organization.

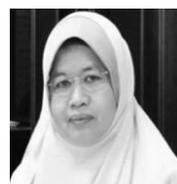

**Dr. Zulaiha Ali Othman** currently an Associate Professor with the Centre of Artificial Intelligence Technology, Faculty of Information Science and Technology, UKM. She is also the Head of the ICT Unit, Centre of Research Innovation and Management, UKM. Since 2003, she has been involved with various intelligent system projects, particular in developing intelligent techniques based on artificial intelligence for problem-solving agents, knowledge discovery, searching, data analytic, and knowledge manipulation. She has vast experience in framework development, algorithm development, and applied artificial intelligence solutions in various domain problems, such as network intrusion detection, human talent, poverty, and weather and air pollution. She has conducted many local, industry, and an international project, which totals more up to RM ten million. She also had graduated with more than 20 Ph.D. students who come from around the world. Besides academics, she is very concerned about community development. She involves in various NGO activities helping the people are needed in a regular basis. She has published more than 200 articles in various local and international publications, includes the high impact journal as well as Journal of an Expert System with Application, Applied Intelligence, Intelligent Data Analysis, Applied Soft Computing, and so on.